Textual Paralanguage and Its Implications for Marketing Communications[†]

Andrea Webb Luangrath[a]

University of Iowa

Joann Peck

University of Wisconsin–Madison

Victor A. Barger

University of Wisconsin–Whitewater

[a] Andrea Webb Luangrath (awluangrath@gmail.com) is an Assistant Professor of Marketing at the University of Iowa, 21 E. Market St., Iowa City, IA 52242. Joann Peck (joann.peck@wisc.edu) is an Associate Professor of Marketing at the University of Wisconsin–Madison, 975 University Ave., Madison, WI 53706. Victor A. Barger (bargerv@uww.edu) is an Assistant Professor of Marketing at the University of Wisconsin–Whitewater, 800 W. Main St., Whitewater, WI 53190. The authors would like to thank the editor, associate editor, and three anonymous reviewers for their helpful feedback, as well as Veronica Brozyna, Laura Schoenike, and Tessa Strack for their research assistance.

## Abstract

Both face-to-face communication and communication in online environments convey information beyond the actual verbal message. In a traditional face-to-face conversation, paralanguage, or the ancillary meaning- and emotion-laden aspects of speech that are not actual verbal prose, gives contextual information that allows interactors to more appropriately understand the message being conveyed. In this paper, we conceptualize textual paralanguage (TPL), which we define as written manifestations of nonverbal audible, tactile, and visual elements that supplement or replace written language and that can be expressed through words, symbols, images, punctuation, demarcations, or any combination of these elements. We develop a typology of textual paralanguage using data from Twitter, Facebook, and Instagram. We present a conceptual framework of antecedents and consequences of brands' use of textual paralanguage. Implications for theory and practice are discussed.

A customer of Whole Foods tweets that he received a bad cupcake from the grocer, to which Whole Foods replies, “A bad cupcake?!!?! Oh No!!! I’m so sorry. *sigh* Thank you for letting us know” (Whole Foods Market, 2013). How does communication on social media affect brand perceptions? Marketers are communicating with customers using a “shorthand, digital language” (Smith, 2015), yet the nature of these communications is under-investigated.

In marketing, research on linguistics has focused primarily on the effects of word choice, such as the effect of explanatory words on consumption experiences (Moore, 2012), refusal words on choice (Patrick & Hagtvedt, 2012), and vowel sounds in brand names on brand preferences (Lowrey & Shrum, 2007). We also see evidence that imperative messages (e.g., “Buy Now!”) are more effective in uncommitted consumer-brand relationships (Moore, Zemack-Rugar, & Fitzsimons, working paper), and assertive statements are more effective at garnering consumer compliance for hedonic products (Kronrod, Grinstein, & Wathieu, 2012). In contrast, our work focuses not on the words said, but on the way nonverbal information is conveyed in writing.

As computer-mediated communication (CMC) has become more prevalent, people have evolved new ways of communicating. Electronic messages are often imbued with nonverbal cues that signal individual characteristics, attitudes, and emotions. Indeed, various researchers recognize that people adapt to the limitations of CMC by creating surrogates for missing social cues (Byron & Baldridge, 2007; Ganster, Eimler, & Krämer, 2012; Walther, 1996). The primary goal of this paper is to provide a framework for the surrogates that people are using in digital communications.

We define *textual paralanguage* (TPL) as *written manifestations of nonverbal audible, tactile, and visual elements that supplement or replace written language and that can be* *expressed through words, symbols, images, punctuation, demarcations, or any combination of these elements*. Expression of nonverbals in text typically differs from the verbal message in several ways: (1) the words are delineated by special characters (e.g., "*") or styles (e.g., CAPS), (2) the words are not standard English but still possess meaning, (3) the words do not flow grammatically with the sentence, and/or (4) the nonverbals occur in the form of a visual image (e.g., emoji). The Whole Foods' tweet, for example, contains four instances of TPL: "?!!?!", "Oh", "!!!", and "*sigh*".

In this paper, we take both an inductive and a deductive approach to the conceptualization of TPL, first exploring how linguistic theory informs the study of TPL, then analyzing how companies are using TPL in their online interactions. We theorize five types of TPL and conclude with a discussion of theoretical and managerial implications as well as avenues for future research.

## In-Person Nonverbal Communication and Behavior

Nonverbal communication refers to communication effected by means other than words (Knapp, Hall, & Horgan, 2013). It is readily observed in all in-person interactions, yet the notion of what constitutes nonverbal communication online is not as clear. To understand the nature of nonverbals in text, we first explore nonverbals in face-to-face interactions.

### Auditory Nonverbal Communication

One of the earliest theorists to study nonverbal communication was Trager (1958, 1960), who noted the depth and importance of information communicated by aspects of speech such as pitch, rhythm, and tempo. Trager (1958) described paralanguage in terms of vocal qualities and vocalizations that qualify literal words. These vocal properties have been termed "implicit" aspects of speech (Mehrabian, 1970) since human speech is naturally imbued with vocal sounds. Communicating aspects of speech aside from literal words has been common among playwrights for centuries. In cinema and theater, paralinguistic elements are inserted into scripts to give stage directions, relay emotions, and facilitate interaction, guiding theatrical performance across languages, cultures, and time (Poyatos, 2008).

## Visual Nonverbal Communication

Just as auditory aspects of speech are inherent in face-to-face communication, so too are visual elements of communication. Birdwhistell (1970) investigated kinesics, the conscious or unconscious bodily movements that possess communicative value, including human gestures and body language. An important bodily communicator is the human face; some scholars claim that it is the primary source of communicative information next to human speech (Knapp et al., 2013). Subtle changes in facial muscle movements can communicate emotional states and provide nonverbal feedback (Ekman et al., 1987). It is thus not surprising that visual textual paralinguistic elements exist in the form of facial emojis.

Nonverbal visual elements are not exclusively related to bodily movements. Visual presentational style conveys information in face-to-face communication through adornments, clothing, style, tattoos, and cosmetics (Barnard, 2001). Often referred to as artifacts, these stylistic choices possess nonverbal signaling power that can communicate personality characteristics (Back, Schmulke, & Egloff, 2010) and are often the basis for initial judgments and impressions.

## Haptic Nonverbal Communication

Touch is the most basic form of communication; indeed, at birth the sense of touch is the most developed of our senses (Hall, 1966; Knapp et al., 2013). Young children use touch to explore their environment, and later in life touch becomes an effective method for communicating with others. We know that individuals have differing preferences for touch in interactions with others, with some people seeking out touch when others avoid it (Webb & Peck, 2015). The meaning of touch in interaction is highly dependent on environmental, personal, and contextual factors. Recent research shows that the degree of relationship closeness influences the types of touch that are deemed appropriate (Suvilehto, Glerean, Dunbar, Hari, & Nummenmaa, 2015).

## Nonverbal Communication Online and Textual Paralanguage Conceptualization

Given the importance of nonverbal communication in face-to-face interactions, it is reasonable to assume that nonverbals play an important role in textual communication as well. Various researchers have noted the presence of paralinguistic elements in text-based messages (e.g., Lea & Spears, 1992; Poyatos, 2008). Lea and Spears (1992) suggest that paralinguistic marks, which they define as ellipses, inverted commas, quotation marks, and exclamation marks, affect perceptions of anonymous communicators online. Although symbols and punctuation possess communicative value, a broader conceptualization of textual paralanguage is needed. To this end, we propose a typology for categorizing and differentiating the various paralinguistic elements that occur in text. It is our hope that this typology will facilitate future research on TPL, its antecedents, and its consequences.

Combining theoretical perspectives on verbal and nonverbal communication, we assert that in-person paralanguage and text-based paralanguage vary in three consequential ways. First, face-to-face paralanguage is typically superimposed on the message, whereas TPL is often decomposed. That is, in face-to-face communication, the verbal and nonverbal elements are combined; vocal aspects of speech are inherent in the production of speech, and gestures occur concurrently with the message (Key, 1975). In text-based communication, however, it is possible for the paralinguistic element (e.g., *wink*) to occur before or after the verbal component of the message.

Second, paralanguage in face-to-face communication is more likely to be processed nonconsciously; that is, in-person gestures and nonverbals are encoded and decoded with varying degrees of awareness and control (Knapp et al., 2013). In text, however, encoding and decoding of paralanguage is more likely to be a conscious process. Whereas in-person nonverbals may be incidental or automatically enacted (e.g., smiling while talking), nonverbals in text tend to be more deliberate and intentional (e.g., inserting a smiley face).

Third, when communicating in-person, paralanguage may be seen, heard, or felt, but in text it is visual, since it is through the eyes that the message and accompanying paralanguage are received. Although audible and haptic cues are referenced in text, no auditory or haptic stimuli are experienced. That said, TPL may evoke imagery of represented gestures, sounds, or facial expressions, which can make the message more concrete and realistic (Borst & Kosslyn, 2010).

Our typology of TPL (figure 1) is based on the senses predominantly used in human interaction: sound, touch, and visuals, rather than taste and smell, which are more relevant for personal experience. From the literature, we identified auditory, tactile, and visual properties of communication that are likely to occur in text. Consistent with previous research on paralanguage, we distinguish between voice qualities, vocalizations, and kinesics (Key, 1975). We further add a category of "artifacts" to accommodate visuals in text that may not correspond directly to in-person communication. We elaborate on each of these in the following paragraphs.

**Voice Qualities.** *Voice qualities are characteristics of the sound of the words being communicated that have to do with how the word(s) should be spoken.* This type of paralanguage represents auditory properties and incorporates aspects such as emphasis, pitch, and rhythm. Voice qualities are often communicated through capitalization, underlining, punctuation, and special characters (e.g., an asterisk). An example of a message that conveys voice qualities, and more specifically rhythm, is "Best. Sale. Ever." The rhythm of the message is indicated by the periods after each word. Thus, the TPL imbues the message with additional significance, and "Best. Sale. Ever." conveys more information than "Best sale ever." There are also non-standard spellings of words that are intentionally written to convey sound qualities. As Carey (1980, p. 67) notes, "[mis]spelling may serve to mark a regional accent or an idiosyncratic manner of speech." For example, "vell vell" suggests a different intonation than "well well".

**Vocalizations.** *Vocalizations are utterances, fillers, terms, or sounds that can be spoken or produced by the body and that result in an audible noise that is recognizable*. Vocalizations are not necessarily English words, but they do convey meaning. Examples include utterances such as "umm" or "uhhh," which, depending on the context of the message, may convey hesitancy, nervousness, or indecision. Physiologic or bodily sounds, such as burping or sneezing, are also included in this type of paralanguage. While some vocalizations are clearly not "English words," there are vocal sounds that have been granted "word" status by dictionaries. For example, "uh" and "uh-huh" are considered words by Merriam-Webster. Conversely, "zzz" is not recognized by Merriam-Webster or the Oxford English Dictionary (OED, 2015), although it is found in almost every online dictionary (e.g., Dictionary.com, 2015).

**Tactile Kinesics.** *Tactile kinesics is the conveyance of nonverbal communication related to physical, haptic interaction with another individual.* Tactile kinesic TPL includes interactional elements between two communicating parties through the use of interpersonal touch. For example, "*high five*" is a tactile kinesic because it is suggestive of physical contact between the sender and the recipient.

**Visual Kinesics.** *Visual kinesics is the conveyance of nonverbal communication related to representation or movement of any part of the body or the body as a whole.* Visual kinesics in TPL includes emoticons and emojis that signify bodily movements. Although various researchers have investigated the use of emoticons in online communications (e.g., Kim & Gupta, 2012, Walther & D'Addario, 2001), within our conceptualization emoticons are simply one example of visual kinesic paralanguage. For example, "*eyeroll*" indicates a bodily movement and thus is an example of visual kinesic TPL.

**Artifacts.** *Artifacts are the presentational style of the text-based message*. In text, artifacts pertain to how the message appears: typeface, stylistic spacing, color, formatting, and layout. Investigating written communication in print advertising, Childers and Jass (2002) demonstrate that typeface semantic cues affect brand perceptions. Also included in this category are non-kinesic and non-tactile emojis and stickers, such as the emoji for a car. Images and icons often supplement or replace words in online communications.

[INSERT FIGURE 1]

## Exploratory Study: Brands' Use of Textual Paralanguage

Heretofore we have employed an inductive approach to understanding the TPL phenomenon. In this study we approach TPL deductively; that is, we examine evidentiary data to see how TPL is being used in actual online communications. We examine brand posts on various social media platforms to substantiate our framework.

**Sample**

To adequately capture the TPL phenomenon, we selected large national brands that have a diverse social media presence. It is common for brand communications to originate from both a corporate account (e.g., @Geico) and a spokescharacter account (e.g., @TheGEICOGecko) (Cohen, 2014). For each brand and spokescharacter, the most recent posts from Twitter, Facebook, and Instagram were collected. These text-based messages were then imported into TAMS Analyzer, an open source tool for coding text, and three individuals manually coded the tweets for TPL. (For additional methodological information and analyses, see the Methodological Details Appendix.)

**Results**

In our sample, 20.6% of brand tweets, 19.1% of Facebook posts, and 31.3% of Instagram posts contained TPL. Across the three platforms, there is evidence that all five types of TPL are utilized by brands, with voice qualities appearing most frequently and tactile kinesics least frequently (tables 1, 2, 3 and 4).

Uses of TPL emerged from the data that were not initially theorized from our review of the literature. One example is the spelling out of words. In a Facebook post, Chester the Cheetah (2014) wrote, “How do you spell Flamin’ Hot CHEETOS Burrito? M-I-N-E”. The use of the dashes to separate the letters in “mine” indicates that each *letter* is to be vocalized, thus representing a new instance of voice quality.

[INSERT TABLES 1, 2, 3, 4]

## Antecedents of Textual Paralanguage Use

We now touch on brand, platform, and target audience factors that motivate the use of TPL (figure 2). In online communications, brands try to foster a strong "social presence" and the perception of being "real" (Sung & Mayer, 2012; Tu, 2002). Successful interaction with customers online has been attributed to whether or not an organization can demonstrate a "conversational human voice" (Kelleher, 2009). Many individuals within an organization contribute to the voice of the organization, and the degree to which interactions are interactive, candid, and "human" can have a lasting impact on relational outcomes, especially when encountering negative electronic word of mouth (Van Noort & Willemsen, 2012). Since nonverbal cues are lacking in electronic communication (Walther, 1993), online communicators use TPL to convey meaning and emotion.

Certain product categories, such as orange juice, possess inherent personality differences (e.g., warmth) compared to other product categories, like pain relievers (Bennett & Hill, 2012). TPL may be beneficial for brands that are motivated to create a young, relatable, or warm image. Brands may also choose to use TPL differentially across their communication portfolios. Consumer brands, like people, are imbued with personality traits (Aaker, 1997; Fournier, 1998), often through techniques such as anthropomorphism (Aggarwal & McGill, 2012) and the use of a brand mascot (Brown, 2010), and these characters may be more likely to use TPL. Additionally, the type of TPL employed may depend on the personality of the communicator. Barbe and Milone (1980) identify visual, auditory, and kinesthetic cognitive learning styles. A visual individual may use more artifacts, a kinesthetic communicator may prefer tactile kinesics, and an auditory-oriented individual may favor vocalizations.

Besides brand considerations, platform-specific norms of communication may guide the use of TPL. For example, the character limit on Twitter encourages posters to find unique ways of constructing messages to save space (e.g., ☺). In addition, platforms are characterized by differences in synchronicity (Porter, 2004). In synchronous communication, conversations take place in real time through written language (Hoffman & Novak, 1996), as in online chats with customer service representatives. In asynchronous communication, posting, viewing, and responding takes place at intervals of time. Since synchronous communication requires immediate responses, message length is necessarily limited, and it is possible that synchronous interactions will contain more TPL.

Communications also vary based on the target or the intended recipient of a message. For example, a younger target may respond more positively to the informal nature of TPL. When a brand is communicating directly with one customer, the personality of the recipient is likely to influence whether TPL is used and how it is interpreted. If a brand is interacting with an expressive and emotional consumer, more consideration may be given to the use of TPL.

## Consequences of Textual Paralanguage Use

TPL has potential downstream consequences for brands (figure 2). For example, TPL is likely to impact perceptions of a brand's personality (Aaker, 1997). Warmth and competence are two characteristics that brands may cultivate, since these translate into increased consumer engagement, connection, and loyalty (Aaker, Garbinsky, & Vohs, 2012). Emoticons, for example, are used more in communications with friends than strangers (Derks, Bos, & Von Grumbkow, 2008) and may foster feelings of warmth and personableness. Emoticons have also been viewed as casual and unprofessional (Jett, 2005), though, and the level of informality associated with TPL could potentially hurt perceptions of firm competence.

Aside from perceptions of a brand's personality, TPL has the potential to influence the brand-consumer relationship. Tactile kinesics, for example, may be used to convey relationship closeness. Many of the textual paralinguistic elements that fall into this category are of a personal nature (e.g., "*hug*"), which foster a sense of closeness.

On the consumer end, TPL may affect message interpretation. Derks et al. (2008) show that emoticons strengthen the intensity of a message. They find that emoticons often serve the same functions as nonverbal behavior and aid in message comprehension. Brand and consumer effects of TPL remain unstudied empirically, and in the next section we consider avenues for future research.

[INSERT FIGURE 2]

## General Discussion and Future Research

In 2015 the Oxford Dictionaries chose, for the first time ever, an emoji as the word of the year (Dictionaries, 2015). Textual paralanguage has become germane to consumer and marketing communications, and it carries the potential to shape how messages are understood. This work suggests that there exists much complexity in the way in which textual messages are used and interpreted. By developing a typology of TPL, we have attempted to make it easier for future researchers to study the properties of text and their various effects on marketing communications.

The TPL dictionary is infinite and ever-expanding. From an etymological perspective, the number of words (and symbols) that we use to communicate meaning has grown exponentially with CMC. It is important to note that nonverbal cues, like verbal ones, rarely have a single denotative meaning; rather, meaning depends greatly on the social context in which the communication resides. Furthermore, the categories of TPL are generally, although not absolutely, mutually exclusive. For example, "*sigh*" can be interpreted as the sound of breath being exhaled forcefully (vocalization), or as the bodily movements associated with sighing, such as shrugging one's shoulders forward or physically looking down (visual kinesics). Notwithstanding examples like this, most instances of TPL are readily classifiable.

Various scholars acknowledge the need for more research on language in consumer psychology (e.g., Kronrod & Danziger, 2013; Schellekens, Verlegh, & Smidts, 2010; Sela, Wheeler, & Sarial-Abi, 2012). Krishna (2012) calls for work on the extent to which language comprehension is bodily grounded. "Can a product description make something smell, feel, sound different? There is an enormous need for research exploring the effect of verbal information on sensory perception" (Krishna, 2012, p. 347). Similarly, can the use of TPL alter sensory experiences? Our TPL typology provides the foundation for exploring these questions.

Auditory, tactile, and visual TPL may be processed differently. There is evidence that modality influences how attitudes are formed, remembered, and altered. Tavassoli and Fitzsimons (2006) demonstrate that attitudes expressed through oral and written communication recruit different cognitive, motor, and perceptual systems and result in the encoding of differentiated memory traces. When the same information is presented in varied contexts, multiple routes are formed in memory. Ease of encoding and response latencies in decoding the types of TPL might differ across individuals' auditory, tactile, and visual learning styles. Future research should consider how the types of TPL are encoded in memory and how this affects retrieval and use of information.

Mental imagery relies on sensory experiences represented in working memory (MacInnis & Price, 1987), and TPL is likely to evoke strong auditory, haptic, and visual imagery. We anticipate that the different types of TPL evoke imagery corresponding to the sensory experience being conveyed, but we also know that imagery systems are interrelated, for example haptic and visual imagery can occur simultaneously (Peck, Barger, & Webb, 2013). There are also individual differences in both the ease of processing and the vividness of imagery (Childers, Houston, & Heckler, 1985). The exploration of imagery evoked by TPL thus promises to be an intriguing area of research.

Nonverbal communication may be processed by either hemisphere of the brain, although the left hemisphere is thought to process more of the verbal and linguistic aspects of communication, and the right hemisphere is credited with visual/spatial relationships, Gestalt information, and the bulk of nonverbal information (Knapp et al., 2013). It would be interesting to test if visual and alphabetic TPL are processed in different regions of the brain. Perhaps characteristics of communicators, such as left vs. right brain dominance, affect the types of TPL they employ. For example, right-brain dominance may lead to more image-based TPL (e.g., emojis), whereas left-brain dominance may favor TPL that modifies words (e.g., loooooong).

If a consumer employs TPL while interacting with a customer service representative, does mimicry of the consumer's writing style by the representative affect what the consumer thinks of the service? We would expect so. Previous research shows that language accommodation is important for customer satisfaction (Van Vaerenbergh & Holmqvist, 2013).

Concordance or discordance in the use of TPL in conversation may affect the way a consumer perceives a brand.

Relatedly, physical mimicry could be investigated. When a consumer is reading a message that contains TPL, does she unconsciously simulate or mimic the expression? For instance, when encountering "*shrug*", do people physically shrug their shoulders? There is research to suggest that when reading auditory cues, people sound out words or imitate how they believe the words to be communicated (Ehri, 2005). We know that when we form perceptions, it is not just a cognitive process, but also an emotional (Loewenstein, 2000) and physiologic (Barsalou, 2008; Carney, Cuddy, & Yap, 2010) one.

There is evidence that language is embodied as well. A growing literature on linguistic embodiment suggests that comprehension relies on internal simulation and bodily action (Fischer & Zwaan, 2008). Recent research on phonetic embodiment finds that phonetic structure influences meaning, as in the direction of tongue movement influencing approach-avoidance tendencies (Topolinski, Maschmann, Pecher, & Winkielman, 2014) and perceptions of acceptance or rejection of a brand name (Kronrod, Lowrey, & Ackerman, working paper). Linking TPL that employs embodiment to measures such as recall and recognition would be a promising area of study.

Conceptually, this research has focused on brands' use of TPL in communications with consumers. However, future research could explore what companies can understand about consumers based on their personal usage of TPL. Can we predict personality, loyalty, or engagement based on TPL? Language use is an individual difference and a meaningful way of exploring personality (Pennebaker & King, 1999). TPL could be used as a predictor of customer personality, tendencies, and behaviors, including age, gender, socioeconomic status, education level, emotional intelligence, closeness of relationships, structure of networks, sentiment, and purchase behavior.

From a managerial perspective, TPL is an important consideration when connecting with consumers online. Choosing whom to hire to manage a brand's social presence has an immense impact on the personality of the brand. Hiring and training decisions should consider TPL, which is a facet of one's tone and "voice" in online communication. For example, a customer service representative who uses online chat to address consumer complaints may need to utilize different communication strategies depending on the source, valence, and context of the message.

Online communication has qualities of both spoken and written language, but it is truly neither. Although early work on interactional and conversational research in marketing acknowledges that nonverbal factors have an immense impact on the interpretation of a marketing message, it was thought that "paralanguage can be eliminated only in situations in which stimulus materials are presented in the form of written dialogue" (Thomas, 1992, p. 89). It is possible for written content to be devoid of paralanguage, but this is rarely the case. Paralanguage is abundant in online communication, and its use will continue to grow with social media. Language, as the basis for human interaction (Grice, 1975), has the capacity to reveal our social and psychological selves. Textual paralanguage contains a wealth of information that marketers should be eager to explore.

## Figure 1. Typology of Textual Paralanguage (TPL)

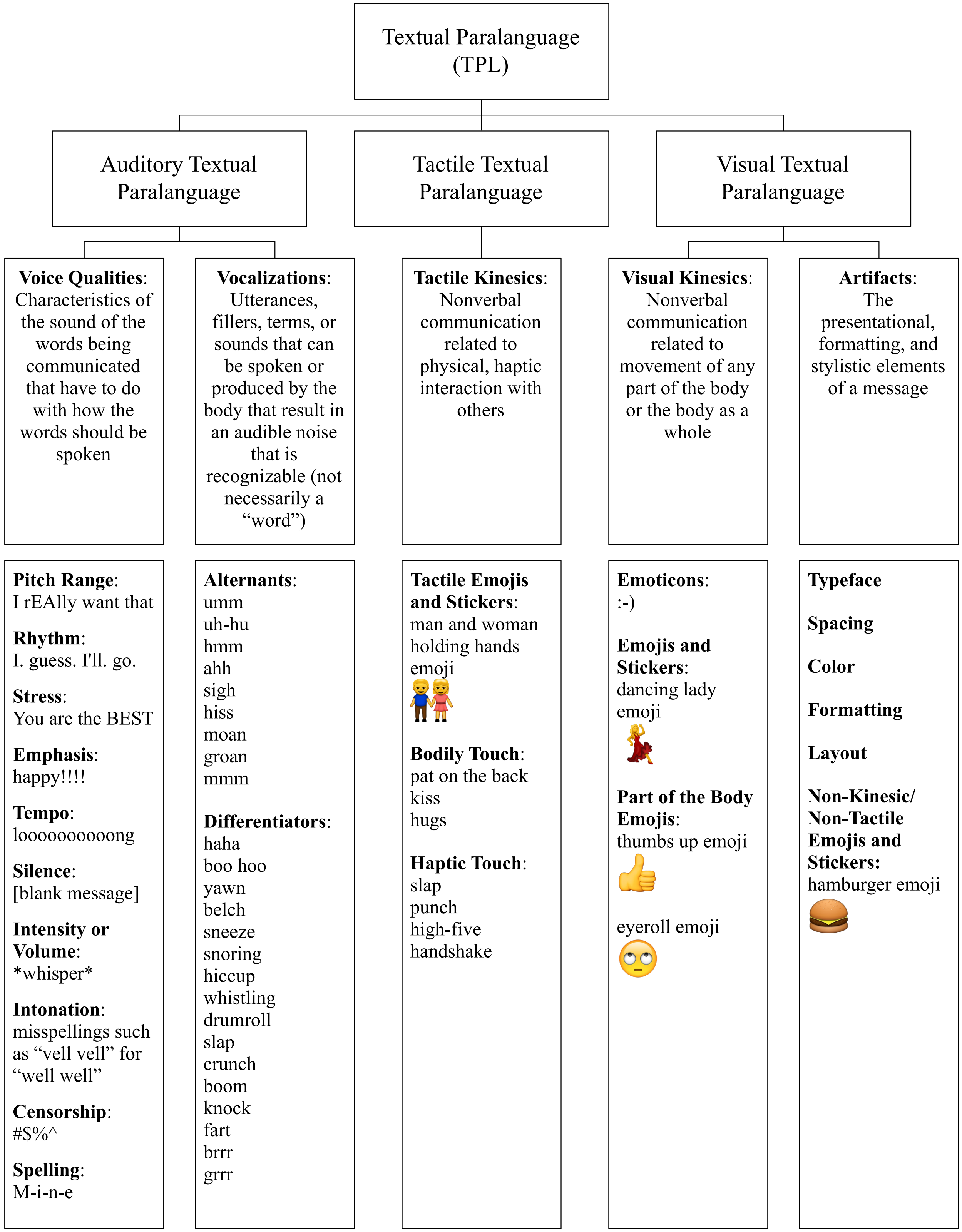

**Figure 2. Conceptual Framework of Antecedents and Consequences of Brands' Use of TPL**

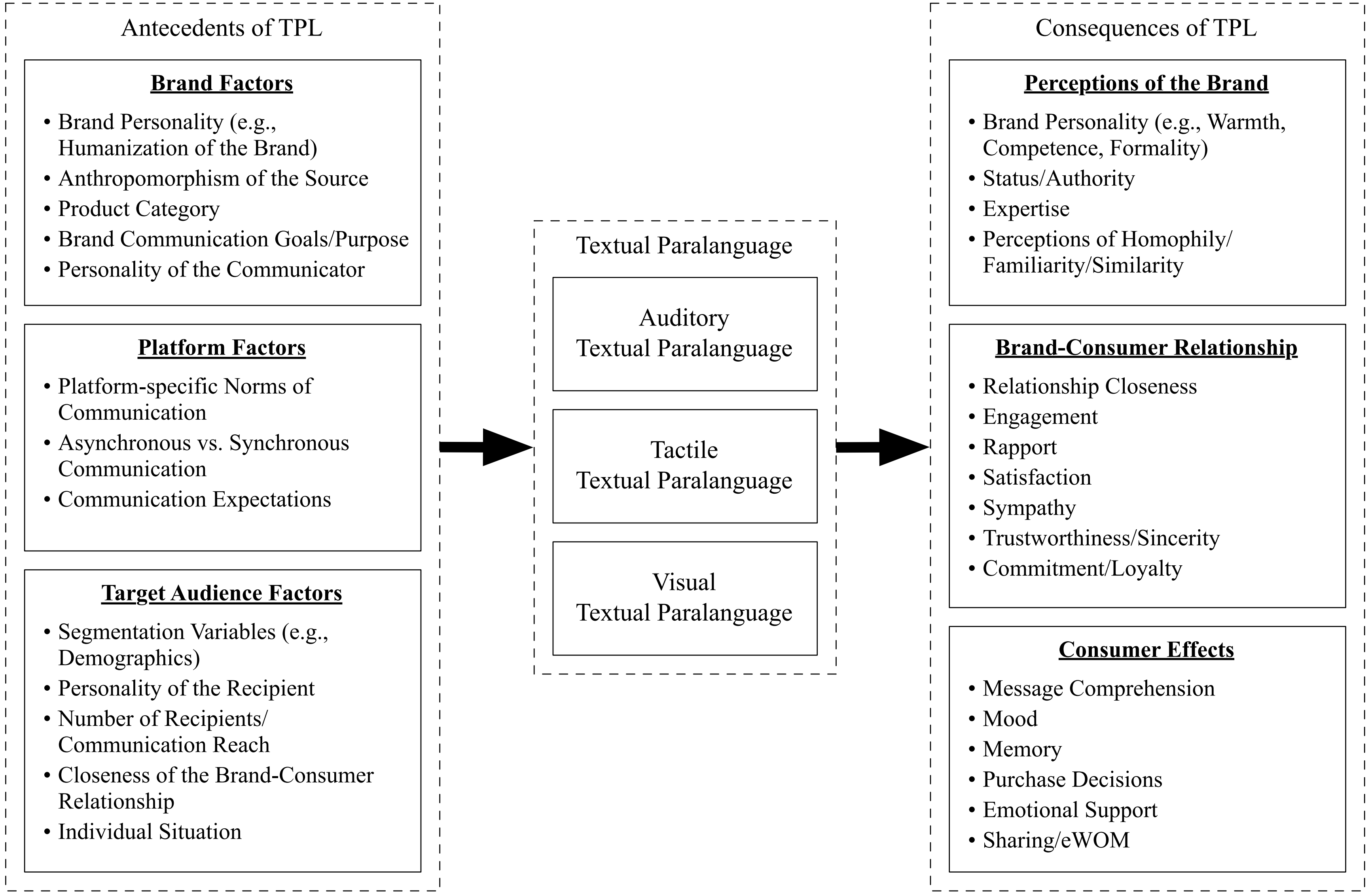

**Table 1: Types of Textual Paralanguage Used by Brands on Twitter**

| Account Type | Twitter Handle | Instances of TPL | Voice Quality | Vocalization | Tactile Kinesic | Visual Kinesic | Artifact |
|---|---|---|---|---|---|---|---|
| **Corporate** | aflac | 3 | 3 (100.0%) | 0 (0.0%) | 0 (0.0%) | 0 (0.0%) | 0 (0.0%) |
| | cheerios | 58 | 30 (51.7%) | 5 (8.6%) | 0 (0.0%) | 12 (20.7%) | 11 (19.0%) |
| | energizer | 11 | 10 (90.9%) | 1 (9.1%) | 0 (0.0%) | 0 (0.0%) | 0 (0.0%) |
| | forestservice | 25 | 20 (80.0%) | 5 (20.0%) | 0 (0.0%) | 0 (0.0%) | 0 (0.0%) |
| | fritolay | 39 | 27 (69.2%) | 7 (17.9%) | 0 (0.0%) | 2 (5.1%) | 3 (7.7%) |
| | geico | 65 | 55 (84.6%) | 4 (6.2%) | 0 (0.0%) | 4 (6.2%) | 2 (3.1%) |
| | kelloggsus | 40 | 27 (67.5%) | 2 (5.0%) | 0 (0.0%) | 5 (12.5%) | 6 (15.0%) |
| | progressive | 21 | 16 (76.2%) | 4 (19.0%) | 0 (0.0%) | 0 (0.0%) | 1 (4.8%) |
| | starbucks | 57 | 19 (33.3%) | 8 (14.0%) | 0 (0.0%) | 9 (15.8%) | 21 (36.8%) |
| | tootsieroll | 85 | 39 (45.9%) | 7 (8.2%) | 0 (0.0%) | 30 (35.3%) | 9 (10.6%) |
| | | | **60.9%** | **10.6%** | **0.0%** | **15.3%** | **13.1%** |
| **Spokescharacter** | aflacduck | 38 | 26 (68.4%) | 7 (18.4%) | 0 (0.0%) | 3 (7.9%) | 2 (5.3%) |
| | buzzthebee | 82 | 48 (58.5%) | 19 (23.2%) | 0 (0.0%) | 8 (9.8%) | 7 (8.5%) |
| | chestercheetah | 41 | 23 (56.1%) | 4 (9.8%) | 0 (0.0%) | 7 (17.1%) | 7 (17.1%) |
| | energizerbunny | 26 | 21 (80.8%) | 3 (11.5%) | 0 (0.0%) | 1 (3.8%) | 1 (3.8%) |
| | frappuccino | 280 | 74 (26.4%) | 31 (11.1%) | 3 (1.1%) | 53 (18.9%) | 119 (42.5%) |
| | itsflo | 52 | 37 (71.2%) | 10 (19.2%) | 0 (0.0%) | 3 (5.8%) | 2 (3.8%) |
| | mrowl | 112 | 70 (62.5%) | 12 (10.7%) | 1 (0.9%) | 22 (19.6%) | 7 (6.3%) |
| | realtonytiger | 50 | 44 (88.0%) | 5 (10.0%) | 0 (0.0%) | 1 (2.0%) | 0 (0.0%) |
| | smokey_bear | 59 | 37 (62.7%) | 9 (15.3%) | 1 (1.7%) | 9 (15.3%) | 3 (5.1%) |
| | thegeicogecko | 37 | 30 (81.1%) | 5 (13.5%) | 0 (0.0%) | 1 (2.7%) | 1 (2.7%) |
| | therealpsl | 26 | 12 (46.2%) | 6 (23.1%) | 2 (7.7%) | 5 (19.2%) | 1 (3.8%) |
| | woodsyowl | 26 | 18 (69.2%) | 3 (11.5%) | 0 (0.0%) | 2 (7.7%) | 3 (11.5%) |
| | | | **53.1%** | **13.8%** | **0.8%** | **13.9%** | **18.5%** |
| **Overall** | | | **55.6%** | **12.7%** | **0.6%** | **14.4%** | **16.7%** |

All frequencies and percentages are based on 200 tweets per Twitter handle, with the exception of frappuccino (N=194), starbucks (N=122), and therealpsl (N=52). Of the 4,168 brand tweets that were analyzed, 859 (20.6%) contained one or more instances of TPL.

**Table 2: Types of Textual Paralanguage Used by Brands on Facebook**

| Account Type | Facebook Page | Instances of TPL | Voice Quality | Vocalization | Tactile Kinesic | Visual Kinesic | Artifact |
|---|---|---|---|---|---|---|---|
| **Corporate** | aflac | 35 | 31 (88.6%) | 2 (5.7%) | 0 (0.0%) | 0 (0.0%) | 2 (5.7%) |
| | cheerios | 51 | 31 (60.8%) | 3 (5.9%) | 0 (0.0%) | 8 (15.7%) | 9 (17.6%) |
| | cheetos | 37 | 30 (81.1%) | 5 (13.5%) | 0 (0.0%) | 0 (0.0%) | 2 (5.4%) |
| | energizer | 29 | 28 (96.6%) | 1 (3.4%) | 0 (0.0%) | 0 (0.0%) | 0 (0.0%) |
| | fritolay | 39 | 36 (92.3%) | 1 (2.6%) | 0 (0.0%) | 0 (0.0%) | 2 (5.1%) |
| | geico | 47 | 38 (80.9%) | 6 (12.8%) | 0 (0.0%) | 3 (6.4%) | 0 (0.0%) |
| | kelloggs | 25 | 21 (84.0%) | 4 (16.0%) | 0 (0.0%) | 0 (0.0%) | 0 (0.0%) |
| | progressive | 24 | 20 (83.3%) | 2 (8.3%) | 0 (0.0%) | 1 (4.2%) | 1 (4.2%) |
| | starbucks | 78 | 41 (52.6%) | 3 (3.8%) | 0 (0.0%) | 14 (17.9%) | 20 (25.6%) |
| | tootsieroll | 168 | 64 (38.1%) | 2 (1.2%) | 1 (0.6%) | 97 (57.7%) | 4 (2.4%) |
| | | | **63.8%** | **5.4%** | **0.2%** | **23.1%** | **7.5%** |
| **Spokescharacter** | aflacduck | 72 | 46 (63.9%) | 18 (25.0%) | 0 (0.0%) | 2 (2.8%) | 6 (8.3%) |
| | energizerbunny | 61 | 52 (85.2%) | 4 (6.6%) | 0 (0.0%) | 4 (6.6%) | 1 (1.6%) |
| | frappuccino | 141 | 71 (50.4%) | 23 (16.3%) | 0 (0.0%) | 20 (14.2%) | 27 (19.1%) |
| | smokeybear | 52 | 42 (80.8%) | 6 (11.5%) | 1 (1.9%) | 1 (1.9%) | 2 (3.8%) |
| | thegeicogecko | 74 | 41 (55.4%) | 3 (4.1%) | 0 (0.0%) | 26 (35.1%) | 4 (5.4%) |
| | | | **63.0%** | **13.5%** | **0.3%** | **13.3%** | **10.0%** |
| **Overall** | | | **63.5%** | **8.9%** | **0.2%** | **18.9%** | **8.6%** |

All frequencies and percentages are based on 250 posts per Facebook Page, with the exception of cheerios (N=249). Of the 3,749 Facebook posts that were analyzed, 716 (19.1%) contained one or more instances of TPL.

**Table 3: Types of Textual Paralanguage Used by Brands on Instagram**

| Account Type | Instagram Account | Instances of TPL | Voice Quality | Vocalization | Tactile Kinesic | Visual Kinesic | Artifact |
|---|---|---|---|---|---|---|---|
| **Corporate** | cheerios | 37 | 30 (81.1%) | 5 (13.5%) | 0 (0.0%) | 0 (0.0%) | 2 (5.4%) |
| | cheetos | 29 | 28 (96.6%) | 1 (3.4%) | 0 (0.0%) | 0 (0.0%) | 0 (0.0%) |
| | energizer | 61 | 52 (85.2%) | 4 (6.6%) | 0 (0.0%) | 4 (6.6%) | 1 (1.6%) |
| | fritolay | 39 | 36 (92.3%) | 1 (2.6%) | 0 (0.0%) | 0 (0.0%) | 2 (5.1%) |
| | geico | 47 | 38 (80.9%) | 6 (12.8%) | 0 (0.0%) | 3 (6.4%) | 0 (0.0%) |
| | kelloggsus | 25 | 21 (84.0%) | 4 (16.0%) | 0 (0.0%) | 0 (0.0%) | 0 (0.0%) |
| | progressive | 24 | 20 (83.3%) | 2 (8.3%) | 0 (0.0%) | 1 (4.2%) | 1 (4.2%) |
| | starbucks | 78 | 41 (52.6%) | 3 (3.8%) | 0 (0.0%) | 14 (17.9%) | 20 (25.6%) |
| | tootsierolltri | 168 | 64 (38.1%) | 2 (1.2%) | 1 (0.6%) | 97 (57.7%) | 4 (2.4%) |
| | | | **23.3%** | **3.9%** | **0.8%** | **21.1%** | **51.0%** |
| **Spokescharacter** | aflacduck | 72 | 46 (63.9%) | 18 (25.0%) | 0 (0.0%) | 2 (2.8%) | 6 (8.3%) |
| | buzzthebee | 51 | 31 (60.8%) | 3 (5.9%) | 0 (0.0%) | 8 (15.7%) | 9 (17.6%) |
| | frappuccino | 141 | 71 (50.4%) | 23 (16.3%) | 0 (0.0%) | 20 (14.2%) | 27 (19.1%) |
| | smokeybear | 52 | 42 (80.8%) | 6 (11.5%) | 1 (1.9%) | 1 (1.9%) | 2 (3.8%) |
| | therealpsl | 74 | 41 (55.4%) | 3 (4.1%) | 0 (0.0%) | 26 (35.1%) | 4 (5.4%) |
| | | | **19.1%** | **6.2%** | **1.0%** | **23.5%** | **50.1%** |
| **Overall** | | | **20.9%** | **5.2%** | **0.9%** | **22.5%** | **50.5%** |

All frequencies and percentages are based on 160 Instagram posts, with the exception of buzzthebee (N=37), cheerios (N=34), cheetos (N=2), fritolay (N=140), geico (N=70), smokeybear (N=147), therealpsl (N=36), and toosierolltri (N=124). Of the 1,550 Instagram posts that were analyzed, 485 (31.3%) contained one or more instances of TPL.

**Table 4: Types of Textual Paralanguage Used by Brands Across Platforms (Twitter, Facebook, and Instagram)**

| Type | Name | Instances of TPL | Voice Quality | Vocalization | Tactile Kinesic | Visual Kinesic | Artifact |
|---|---|---|---|---|---|---|---|
| **Corporate** | Aflac | 38 | 34 (89.5%) | 2 (5.3%) | 0 (0.0%) | 0 (0.0%) | 2 (5.3%) |
| | Cheerios | 123 | 62 (50.4%) | 9 (7.3%) | 0 (0.0%) | 23 (18.7%) | 29 (23.6%) |
| | Cheetos | 41 | 23 (56.1%) | 4 (9.8%) | 0 (0.0%) | 7 (17.1%) | 7 (17.1%) |
| | Energizer | 52 | 49 (94.2%) | 2 (3.8%) | 0 (0.0%) | 1 (1.9%) | 0 (0.0%) |
| | Forest Service | 25 | 20 (80.0%) | 5 (20.0%) | 0 (0.0%) | 0 (0.0%) | 0 (0.0%) |
| | Fritolay | 107 | 72 (67.3%) | 10 (9.3%) | 2 (1.9%) | 5 (4.7%) | 18 (16.8%) |
| | Geico | 121 | 100 (82.6%) | 12 (9.9%) | 0 (0.0%) | 7 (5.8%) | 2 (1.7%) |
| | Kelloggs | 98 | 60 (61.2%) | 9 (9.2%) | 0 (0.0%) | 5 (5.1%) | 24 (24.5%) |
| | Progressive | 59 | 45 (76.3%) | 8 (13.6%) | 0 (0.0%) | 1 (1.7%) | 5 (8.5%) |
| | Starbucks | 284 | 74 (26.1%) | 11 (3.9%) | 1 (0.4%) | 56 (19.7%) | 142 (50.0%) |
| | Tootsie Roll | 354 | 124 (35.0%) | 13 (3.7%) | 1 (0.3%) | 163 (46.0%) | 53 (15.0%) |
| | | | **50.9%** | **6.5%** | **0.3%** | **20.6%** | **21.7%** |
| **Spokescharacter** | Aflac Duck | 303 | 85 (28.1%) | 35 (11.6%) | 0 (0.0%) | 61 (20.1%) | 122 (40.3%) |
| | Buzz the Bee | 96 | 57 (59.4%) | 23 (24.0%) | 0 (0.0%) | 8 (8.3%) | 8 (8.3%) |
| | Chester Cheetah | 37 | 30 (81.1%) | 5 (13.5%) | 0 (0.0%) | 0 (0.0%) | 2 (5.4%) |
| | Energizer Bunny | 87 | 73 (83.9%) | 7 (8.0%) | 0 (0.0%) | 5 (5.7%) | 2 (2.3%) |
| | Frappuccino | 658 | 195 (29.6%) | 64 (9.7%) | 8 (1.2%) | 128 (19.5%) | 263 (40.0%) |
| | Flo | 52 | 37 (71.2%) | 10 (19.2%) | 0 (0.0%) | 3 (5.8%) | 2 (3.8%) |
| | Mr. Owl | 112 | 70 (62.5%) | 12 (10.7%) | 1 (0.9%) | 22 (19.6%) | 7 (6.3%) |
| | Real Tony Tiger | 50 | 44 (88.0%) | 5 (10.0%) | 0 (0.0%) | 1 (2.0%) | 0 (0.0%) |
| | Smokey Bear | 147 | 91 (61.9%) | 18 (12.2%) | 2 (1.4%) | 15 (10.2%) | 21 (14.3%) |
| | The Geico Gecko | 111 | 71 (64.0%) | 8 (7.2%) | 0 (0.0%) | 27 (24.3%) | 5 (4.5%) |
| | The Real PSL | 43 | 23 (53.5%) | 10 (23.3%) | 2 (4.7%) | 6 (14.0%) | 2 (4.7%) |
| | Woodsy Owl | 26 | 18 (69.2%) | 3 (11.5%) | 0 (0.0%) | 2 (7.7%) | 3 (11.5%) |
| | | | **46.1%** | **11.6%** | **0.8%** | **16.1%** | **25.4%** |
| **Overall** | | | **48.2%** | **9.4%** | **0.6%** | **18.1%** | **23.8%** |

## Methodological Details Appendix

This appendix provides additional detail on the exploratory study reported in the manuscript. To ensure saturation of the TPL phenomenon, we collected data from consumers as well as brands. We describe the analyses we conducted on consumer tweets, brand tweets, brand at-replies, brand posts on Facebook, and brand posts on Instagram.

### Consumer Tweets

To obtain a sample of public tweets, a Python program was written to collect tweets from Twitter for analysis. Twitter is an ideal social media platform for investigating TPL, since posts are primarily textual, messages are limited to 140 characters, and programmatic access to all public tweets is possible using an application programming interface (API). To obtain a sample of all public tweets written in the English language, the program queried the Twitter Streaming with the parameter "language=en". This was done at different times of the day (during daytime hours in the United States) over the course of several days until 5,000 tweets were acquired. This sample provides consumer-level data on how individuals use TPL.

After each query, the tweets were downloaded in JSON format and saved using UTF-8 encoding to preserve emojis and other symbols. The tweets were then imported into TAMS Analyzer, an open-source research tool, for manual coding of textual paralanguage (see Table A1 for coding guide). The coders were instructed to identify all instances of nonverbal communication in text, regardless of its fit within the existing categories. It was made clear that the purpose was to uncover whether or not the existing classification was indeed the correct one, or whether categories exist that are not captured using the current framework. The tweets were coded independently by four coders, and the resulting documents were compared using Kaleidoscope. Discrepancies were resolved by discussion amongst the researchers.

Of the 5,000 randomly sampled tweets, 4,608 (92.2%) were valid tweets. Tweets were coded as not valid if they used languages other than English (0.1%), if they were generated automatically by a program (3.7%), or were spam (4%). Of the 4,608 valid tweets, 1,859 (40.3%) employed some form of TPL. Clearly *how* messages are written matters. The prevalence of the various types of TPL is important as well. Of the 3,097 instances of TPL, voice quality was the most common (35.4%), with visual kinesics a close second (33.7%). This was followed by artifacts (16.4%), vocalizations (11.5%), and tactile kinesics (3%).

**Brand Tweets**

Twitter is not only used by consumers but is also widely used by brands (King, 2008). A Python program was written to collect brand tweets for analysis. For each brand, the program queried the Twitter REST API, downloaded the tweets in JSON format, and saved the tweets using UTF-8 encoding to preserve emojis and other symbols. All at-replies (tweets that begin with "@") were excluded from this sample, since these are primarily responses to tweets from other Twitter users; in addition, at-replies are typically only seen by the intended recipient of the tweet. For comprehensiveness, however, we analyze at-replies in the following section. Retweets were also excluded, since the text of a retweet is not composed by the brand. The most recent 200 tweets for each brand were imported into TAMS Analyzer for coding. Only three of the brand accounts had fewer than 200 tweets after removing retweets and at-replies: frappuccino (N=194), starbucks (N=122), and therealpsl (N=52). The tweets were coded independently by four coders, and the resulting documents were compared using Kaleidoscope. Discrepancies were resolved by discussion amongst the researchers.

Of the 4,168 brand tweets that were analyzed, 859 (20.6%) contained one or more instances of TPL (see Table 1 in the manuscript). In all there were 1,233 instances of TPL use, of which 55.6% were voice qualities, 12.7% were vocalizations, 0.6% were tactile kinesics, 14.4% were visual kinesics, and 16.7% were artifacts.

**Brand At-Replies**

A Python program was written to collect brand at-replies for analysis. For each brand, the program queried the Twitter REST API, downloaded the at-replies in JSON format, and saved the at-replies using UTF-8 encoding to preserve emojis and other symbols. The most recent 150 at-replies for each brand were imported into TAMS Analyzer for coding. Only five of the brand accounts had fewer than 150 at-replies: aflac (N=29), forestservice (N=7), fritolay (N=149), realtonytiger (N=125), and woodsyowl (N=83). The at-replies were coded independently by four coders, and the resulting documents were compared using Kaleidoscope. Discrepancies were resolved by discussion amongst the researchers.

Of the 2,943 brand at-replies that were analyzed, 1,025 (34.8%) contained one or more instances of TPL (see Table A2). In all there were 1,342 instances of TPL use, of which 25.3% were voice qualities, 21.8% were vocalizations, 2.7% were tactile kinesics, 33.2% were visual kinesics, and 17.1% were artifacts.

**Brand Facebook Posts**

Posts on brand Facebook Pages were downloaded using DiscoverText, a cloud-based text analytics service. After filtering the posts on the Facebook Pages by brand, the most recent 250 brand posts were downloaded and imported into TAMS Analyzer for coding. Only one brand had fewer than 250 posts: cheerios (N=249). The Facebook posts were coded independently by four coders, and the resulting documents were compared using Kaleidoscope. Discrepancies were resolved by discussion amongst the researchers.

Of the 3,749 brand Facebook posts that were analyzed, 716 (19.1%) contained one or more instances of TPL (see Table 2 in the manuscript). In all there were 933 instances of TPL use, of which 63.5% were voice qualities, 8.9% were vocalizations, 0.2% were tactile kinesics, 18.9% were visual kinesics, and 8.6% were artifacts.

**Brand Instagram Posts**

Posts on brand Instagram accounts were downloaded using Iconosquare, a cloud-based service for viewing Instagram posts on the web. After loading each brand's Instagram page on Iconosquare, the page was saved as HTML for scraping with a program written in Python. The most recent 160 Instagram posts for each brand were then imported into TAMS Analyzer for coding. Eight brand accounts had fewer than 160 posts: buzzthebee (N=37), cheerios (N=34), cheetos (N=2), fritolay (N=140), geico (N=70), smokeybear (N=147), therealpsl (N=36), and toosierolltri (N=124). The Instagram posts were coded independently by four coders, and the resulting documents were compared using Kaleidoscope. Discrepancies were resolved by discussion amongst the researchers.

Of the 1,550 brand Instagram posts that were analyzed, 485 (31.3%) contained one or more instances of TPL (see Table 3 in the manuscript). In all there were 858 instances of TPL use, of which 20.9% were voice qualities, 5.2% were vocalizations, 0.9% were tactile kinesics, 22.5% were visual kinesics, and 50.5% were artifacts.

### Table A1: Textual Paralanguage Coding Guide

| | |
|---|---|
| Voice Quality ("VQ") | Denotes how the word(s) should be spoken<br>• Emphasis: really**?!?!?!!** awesome**!!!!**<br>• Stress: You are the **BEST**<br>• Pitch: I **rEAlly** want that<br>• Rhythm: **Best. Day. Ever.** *or* p l e a s e<br>• Tempo: So **looooooooong** *or* I suppose**.....** *or*<br>• Scare quotes: That was **"fun"**.<br>• Silence: [blank messages]<br>• Intensity or Volume: ***whisper***<br>• Intonation:[often communicated through misspellings; e.g., '**vell vell**']<br>• Censorship: **#$%^**<br>• Spelling: **M-i-n-e** |
| Vocal-ization ("VS") | Fillers, meaningful utterances, or bodily sounds (not necessarily a "word")<br>• **aww**<br>• **umm**<br>• **uh, ah, oh**<br>• **huh**<br>• **uh huh**<br>• **grrr**<br>• **BRRR**<br>• **sigh**, ***sigh***, **(sigh)**<br>• **yum, yumyum, mmm**<br>• **yeah**<br>• **yay**<br>• **hmm**<br>• **ahh**<br>• **haha, hehe**<br>• **lmao, lmfao**<br>• **lol, *laughing***, **(laughing)**<br>• **boo hoo**<br>• **woah**<br>• **hmph**<br>• **whew**<br>• **Ewwwww**<br>• **Ouch**<br>• **Oops**<br>• **hiss**<br>• **moan**<br>• **groan**<br>• **drumroll**<br>• **slap**<br>• **knock**<br>• **fart**<br>• **crunch**<br>• **boom**<br>• **yawn**<br>• **belch**<br>• **sneeze**<br>• **snoring**<br>• **hiccup**<br>• **whistling**<br>• **shhh** |
| Tactile Kinesic ("TK") | Nonverbal physical, haptic interaction with others<br>• **xxx** (kisses)<br>• **xoxo**<br>• ***hugs***<br>• **high five**<br>• **fist bump**<br>• **pat on the back**<br>• **slap**<br>• **punch**<br>• **handshake**<br>• stickers/emojis that have to do with touch |
| Visual Kinesic ("VK") | Movement of any part of the body or the body as a whole<br>• **thumbs up** or 👍<br>• **rotfl**<br>• **eyeroll**<br>• **shrug**<br>• **:)** or ☺<br>• **T-T** (crying)<br>• stickers/emojis that are suggestive of the body (including anthropomorphized animal faces) |
| Artifact ("A") | Presentational, formatting, and stylistic elements of a message<br>• **<3**<br>• Color<br>• Typeface<br>• Spacing<br>• Formatting (e.g., lists)<br>• Layout<br>• Non-visual kinesic/non-tactile kinesic emoji |
| bot | Automatically generated tweet (ex: "**I'm at McDonald's 4sq.com/1x53idj**") |
| spam | Text an e-mail program would classify as spam (ex: "**Viagra Cialis cheap! SAVE HERE**") |
| noten | Text in a language other than English |

Additional notes on TPL coding:

- When "…" is used solely to separate tweet content and link, do not code as TPL.
- When "…" is at the end of the tweet because the writer ran out of characters, do not code as TPL.
- A single exclamation point ("!") should not be coded as it is regular punctuation.
- A retweet of spam is still coded as spam, but a retweet of a bot should be coded as a regular tweet, as it is no longer automated (someone actually retweeted those thoughts/ideas/actions/sounds).
- When the same emoji is repeated, it is coded as one element: {A}✨✨✨✨{/A}.
- When different emojis are strung together, they are coded separately: {VK}😄{/VK}{VK}💃{/VK}{TK}😘{/TK}{A}🔥{/A}.
- If there are multiple types of TPL included in one word, code both of the types: {VQ}{VS}hmmmmmmmmmm{/VS}{/VQ}.

**Table A2: Types of Textual Paralanguage Used by Brands in Twitter At-Replies**

| Account Type | Twitter Handle | Instances of TPL | Voice Quality | Vocalization | Tactile Kinesic | Visual Kinesic | Artifact |
|---|---|---|---|---|---|---|---|
| **Corporate** | aflac | 0 | 0 (0.0%) | 0 (0.0%) | 0 (0.0%) | 0 (0.0%) | 0 (0.0%) |
| | cheerios | 69 | 6 (8.7%) | 20 (29.0%) | 0 (0.0%) | 28 (40.6%) | 15 (21.7%) |
| | energizer | 5 | 4 (80.0%) | 0 (0.0%) | 0 (0.0%) | 0 (0.0%) | 1 (20.0%) |
| | forestservice | 5 | 5 (100.0%) | 0 (0.0%) | 0 (0.0%) | 0 (0.0%) | 0 (0.0%) |
| | fritolay | 35 | 0 (0.0%) | 4 (11.4%) | 0 (0.0%) | 31 (88.6%) | 0 (0.0%) |
| | geico | 76 | 14 (18.4%) | 12 (15.8%) | 0 (0.0%) | 45 (59.2%) | 5 (6.6%) |
| | kelloggsus | 41 | 4 (9.8%) | 27 (65.9%) | 0 (0.0%) | 7 (17.1%) | 3 (7.3%) |
| | progressive | 4 | 2 (50.0%) | 1 (25.0%) | 0 (0.0%) | 1 (25.0%) | 0 (0.0%) |
| | starbucks | 129 | 12 (9.3%) | 11 (8.5%) | 1 (0.8%) | 52 (40.3%) | 53 (41.1%) |
| | tootsieroll | 155 | 36 (23.2%) | 63 (40.6%) | 0 (0.0%) | 51 (32.9%) | 5 (3.2%) |
| | | | **16.0%** | **26.6%** | **0.2%** | **41.4%** | **15.8%** |
| **Spokescharacter** | aflacduck | 63 | 12 (19.0%) | 9 (14.3%) | 0 (0.0%) | 7 (11.1%) | 35 (55.6%) |
| | buzzthebee | 103 | 50 (48.5%) | 23 (22.3%) | 3 (2.9%) | 13 (12.6%) | 14 (13.6%) |
| | chestercheetah | 93 | 14 (15.1%) | 15 (16.1%) | 0 (0.0%) | 32 (34.4%) | 32 (34.4%) |
| | energizerbunny | 26 | 5 (19.2%) | 1 (3.8%) | 0 (0.0%) | 20 (76.9%) | 0 (0.0%) |
| | frappuccino | 136 | 26 (19.1%) | 12 (8.8%) | 7 (5.1%) | 36 (26.5%) | 55 (40.4%) |
| | itsflo | 49 | 16 (32.7%) | 10 (20.4%) | 0 (0.0%) | 22 (44.9%) | 1 (2.0%) |
| | mrowl | 75 | 24 (32.0%) | 31 (41.3%) | 0 (0.0%) | 20 (26.7%) | 0 (0.0%) |
| | realtonytiger | 79 | 58 (73.4%) | 13 (16.5%) | 1 (1.3%) | 7 (8.9%) | 0 (0.0%) |
| | smokey_bear | 47 | 9 (19.1%) | 10 (21.3%) | 18 (38.3%) | 10 (21.3%) | 0 (0.0%) |
| | thegeicogecko | 64 | 15 (23.4%) | 10 (15.6%) | 0 (0.0%) | 39 (60.9%) | 0 (0.0%) |
| | therealpsl | 64 | 15 (23.4%) | 16 (25.0%) | 6 (9.4%) | 19 (29.7%) | 8 (12.5%) |
| | woodsyowl | 24 | 13 (54.2%) | 4 (16.7%) | 0 (0.0%) | 5 (20.8%) | 2 (8.3%) |
| | | | **31.2%** | **18.7%** | **4.3%** | **27.9%** | **17.9%** |
| **Overall** | | | **25.3%** | **21.8%** | **2.7%** | **33.2%** | **17.1%** |

All frequencies and percentages are based on 150 at-replies per Twitter handle, with the exception of aflac (N=29), forestservice (N=7), fritolay (N=149), realtonytiger (N=125), and woodsyowl (N=83). Of the 2,943 at-replies that were analyzed, 1,025 (34.8%) contained one or more instances of TPL.